\documentclass[letterpaper]{article} 
\usepackage[]{aaai23}  
\usepackage{times}  
\usepackage{helvet}  
\usepackage{courier}  
\usepackage[hyphens]{url}  
\usepackage{graphicx} 
\def\UrlFont{\rm}  
\usepackage{natbib}  
\usepackage{caption} 
\usepackage{algorithm}
\usepackage{algorithmic}
\usepackage{xspace}

\usepackage{newfloat}
\usepackage{listings}
\DeclareCaptionStyle{ruled}{labelfont=normalfont,labelsep=colon,strut=off} 
\floatstyle{ruled}
\newfloat{listing}{tb}{lst}{}
\floatname{listing}{Listing}
\newcommand{\name}{{\fontfamily{bch}\selectfont{\textsc{Mauve}}}\xspace}

\usepackage{adjustbox}
\usepackage{booktabs}

\usepackage{amsmath}

\title{In BLOOM: Creativity and Affinity in Artificial Lyrics and Art}

\author {
    Evan Crothers,\textsuperscript{\rm 1}
    Herna Viktor,\textsuperscript{\rm 1}
    Nathalie Japkowicz\textsuperscript{\rm 2}
}
\affiliations {
    \textsuperscript{\rm 1} University of Ottawa, Ottawa, Ontario, Canada\\
    \textsuperscript{\rm 2} American University, Washington D.C., U.S.A.\\
    ecrot027@uottawa.ca, hviktor@uottawa.ca, japkowic@american.edu
}

\begin{document}

\maketitle
\begin{abstract}
We apply a large multilingual language model (BLOOM-176B) in open-ended generation of Chinese song lyrics, and evaluate the resulting lyrics for coherence and creativity using human reviewers.  We find that current computational metrics for evaluating large language model outputs (MAUVE) have limitations in evaluation of creative writing.  We note that the human concept of creativity requires lyrics to be both comprehensible and distinctive --- and that humans assess certain types of machine-generated lyrics to score more highly than real lyrics by popular artists.  Inspired by the inherently multimodal nature of album releases, we leverage a Chinese-language stable diffusion model to produce high-quality lyric-guided album art, demonstrating a creative approach for an artist seeking inspiration for an album or single.  Finally, we introduce the MojimLyrics dataset, a Chinese-language dataset of popular song lyrics for future research.
\end{abstract}

\section{Introduction}


Modern music production --- which spans elements including musical composition, songwriting, album artwork, and music videos --- is a  tremendous domain for exploring creativity across modalities.  This work focuses on better understanding subjective human assessment of creative text produced by large language models (LLM), demonstrating that mimicry of an existing sample dataset may not always be the best approach when considering human preferences.


Towards this end, we use BLOOM \cite{BLOOM_gang}, a multilingual LLM with 176B parameters, to generate Chinese song lyrics with varying sampling parameters.  We focus on top-$p$ sampling, as this method has been found to outperform top-$k$ and beam search in prior research \cite{holtzman2019curious}.  In order to assess the quality of the generated lyrics, we task human reviewers to provide feedback on lyrical coherence, creativity, and enjoyment.  Using this data, we demonstrate the relationship between creativity and coherence in human-assessed lyric quality, and highlight differences between subjective human assessment and computational methods such as \name \cite{pillutla2021mauve}.


Beyond comparison of creativity measures, we also dive into new emerging multimodal possibilities related to AI-driven creativity.  Music today is often accompanied by cover art that evokes the aesthetic of the piece.  This art, previously displayed on record covers and lyric booklets, remains prominently displayed on music apps and streaming platforms.  While album cover art may reflect a collection of songs (an album, LP, or EP), digital music publishing allows for easy release of singles, and cover art is increasingly used to accompany individual tracks \cite{leight_2018}.  Using our lyrics, we use the Chinese-language Taiyi stable diffusion model \cite{fengshenbang} to generate an album image inspired by the LLM-generated lyrics.  The demonstrated workflow may be used for inspiration by songwriters seeking a unified lyrical and visual concept for a single or album, as well as being a source of entertainment to music fans.

\section{Related Work}

Previous work has used LLMs to generate classical Chinese poetry \cite{Liao2019GPTbasedGF}.  Work has also been performed which uses GPT-2 as a generation system for various types of creative Chinese writing, contemporary song lyrics among them \cite{Zhang2022QiuNiuAC}.  Related work has used an image-generation network as an intermediate step in generating non-lyrical creative text, based on the idea of an author ``visualizing" while they work \cite{zhu2022visualize}.  In English, research has similarly been done on LLM lyric generation \cite{rodrigues2022lyrics}, including lyrics conditioned on a song's melody \cite{chen2020melody}.

Our work is distinct from previous work in that it focuses on zero-shot lyric generation with a massive multilingual model (BLOOM), is the first to our knowledge to benchmark lyrics against evaluation metrics that strongly correlate to human assessment (\name), and most importantly, presents an investigation into the relationship between computational assessment of open-ended text quality, and subjective measurements.  In this department, we are inspired by previous work on quantifying creativity that found that KL divergence has a strong correlation with human-assessed creativity for word pairs \cite{kuznetsova2013understanding}.

Finally, this work is also the first research work to apply diffusion models as part of a unified approach for generating coordinated cover art in association with song lyrics.  All models and code and from our work will be made available open-source to enable future research\footnote{https://github.com/ecrows/in-bloom}.


\section{Methodology}

We organize our methodology into three major parts: 1) collecting and cleaning Chinese lyrics found online to create a novel dataset of popular Chinese lyrics for use with \name, 2) performing lyric generation effectively with the BLOOM-176B LLM, and 3) utilizing the Taiyi stable diffusion model to generate high-quality art to accompany the lyrics.  


\subsection{Dataset}
\label{ssec:dataset}

For comparison to LLM-generated lyrics, we collect a dataset of song lyrics for popular Chinese songs from the website Mojim.com.  Mojim is a popular lyrics sharing website that focuses on Chinese song lyrics. We request the top popular artists in the ``male", ``female", and ``group" categories, fetch all the albums associated with each artist, and download the lyrics for every song in each album.  Metrics on the dataset can be found in Table \ref{table:mojim}.


Data downloaded from Mojim.com is in raw HTML format and must be cleaned to extract raw lyrics.  This process is summarized as follows.

\begin{itemize}
    \item Removal of HTML tags, as well as the removal of markup symbols not typically part of the lyric text, such as ``*" and ``\#" and various unicode equivalents.
    \item Removal of site-specific text characteristics, such as: a recurring line in Chinese which translates to ``Find more lyrics at Mojim.com", dividing lines composed of hyphens, non-standard labels placed at the start of the song.
    \item Compress instances of many subsequent newline characters into just two newline characters, to reduce excess whitespace which may emerge after removing tags, annotations, and symbol markup.
\end{itemize}

\begin{table}[h]
\centering
\begin{tabular}{lllll}
\toprule
Songs & Artists & Tokens & Size \\\midrule
39747 & 230     & 14,406,854 & 38.4MB  \\\bottomrule
\end{tabular}
\caption{Statistics for the MojimLyrics Dataset}
\label{table:mojim}
\end{table}

\subsection{Lyric Generation}

We generate zero-shot samples using the multilingual BLOOM-176B model \cite{BLOOM_gang} at varying values of $p$ for top-$p$ nucleus sampling \cite{holtzman2019curious}.  By adjusting the probability mass used to sample the next token, we produce varied lyric sequences that can be used to analyze quantitative and subjective text characteristics.  Our evaluation of lyrics focuses on two major approaches:

\begin{itemize}
\item \textbf{Computational Quality}:  We measure computational quality via commonly-used metrics that target language model degeneration (repeated $n$-grams, diversity of tokens, distinct $n$-grams), as well as distributional information via \name \cite{pillutla2021mauve}.  
\item \textbf{Subjective}: Subjective measures of coherence are collected via a carefully survey on Amazon Mechanical Turk.  These measures include an assessment of how creative, coherent, and enjoyable the lyrics are. 
\end{itemize}

Previous research on quantifying creativity indicates that for measuring perceived creativity of word pairs $\{w_1, w_2\}$, the KL divergence of $\{w_1, w_1w_2\}$ is among the most effective measures \cite{kuznetsova2013understanding}.

\name is designed to jointly capture two types of errors between the output distribution $Q$ of a model, and the human distibution $P$.  1) Type I errors, where $Q$ places high mass on text which is unlikely under $P$; and, 2) Type II errors, where Q cannot generate text which is plausible under $P$.  \name does this by summarizing Type I and Type II errors using  Kullback–Leibler (KL) divergences, which are measured softly using a mixture distribution. Due to the similarity between \name's computation and previous work on creativity quantification, it is possible that \name may partially reflect subjective human creativity assessment, which may contribute to \name's strong performance for evaluating generated text quality more generally.


Additional text quality metrics for assessing language model output are selected based on previous work \cite{zhu2022visualize, su2022contrastive}.  These metrics include rep-$n$ = 1.0 - $\frac{|\text{unique }n\text{-grams}|}{|\text{total }n\text{-grams}|}$ measuring duplicate $n$-grams within each generated sequence \cite{Welleck2020Neural},
distinct-$n$ = $\frac{|\text{unique }n\text{-grams}|}{|\text{length of text}|}$ \cite{li2016diversity},
and diversity = $\prod_{n\text{=2}}^{4}(1-\text{rep-}n)$ for measuring diversity of $n$-grams \citep{su2022contrastive}.






Autoregressive language models such as BLOOM perform sampling to select a next token typically using some variation of top-$k$, beam search, and nucleus top-$p$ sampling. Due to the strong documented performance of top-$p$ sampling \cite{holtzman2019curious}, we produce BLOOM generations at varying levels of $p\in[0.80,0.85,0.90,0.95,0.99]$.

\begin{figure}[h]
\centering
\includegraphics[width=0.70\linewidth]{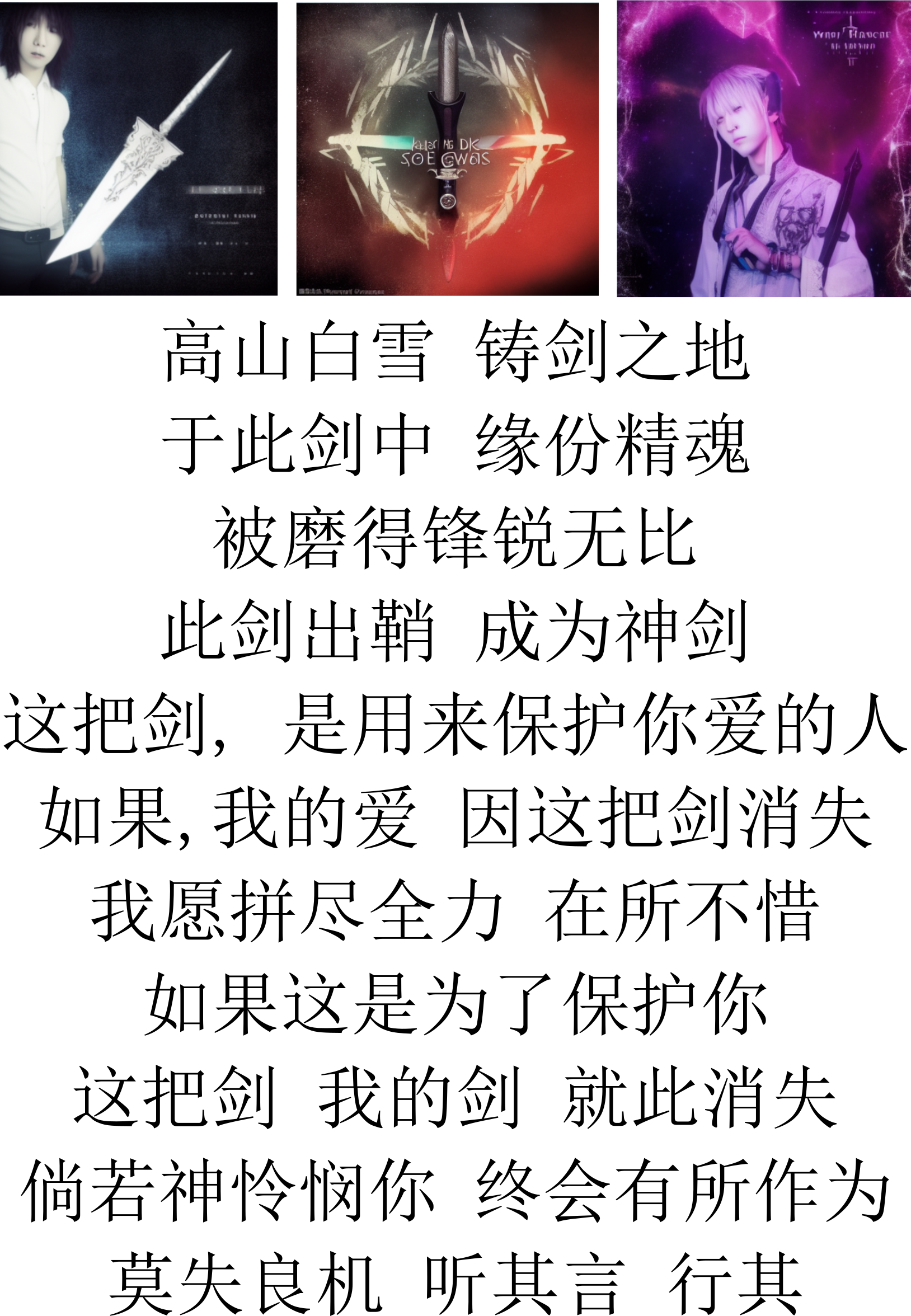}
\caption{BLOOM-176B $p=0.95$, $seed=0$ lyrics mentioning a sword, accompanied by corresponding stable diffusion generated images selected from the resulting prompt at $CFG=0.7$, $seed=0$, $batch\_size=6$}
\label{fig:swordartonlinedieinthegamedieinreallifekekits418am}
\end{figure}

\subsection{Album Art Generation}

We leverage our generated song lyrics to create accompanying track art inspired by the song lyrics.  As a generative model, we utilize the Chinese-language Taiyi Stable Diffusion model \cite{fengshenbang, rombach2021highresolution}, which includes a Chinese-language text encoder.  We use this model to produce 1000 samples of album artwork, from seeds 0-999, using the prompt ``album art" in Chinese, followed by our generated lyrics.  We perform this at varying settings of classifier-free diffusion guidance (CFG), where $CFG\in[4.0,7.0,10.0]$.  Images are generated with 20 sampling steps using the Euler A method \cite{rombach2021highresolution}.  For evaluation, we utilize a dataset of approximately 4,000 album covers at 512x512 resolution \cite{surma_2019}.  We determine the quality of the generated images by calculating Fréchet Inception Distance (FID) \cite{DBLP:journals/corr/HeuselRUNKH17, Seitzer2020FID} using the pool 3 activations from an Inception network \cite{szegedy2015going} between generated album covers and the real album dataset.






\begin{figure}[h]
    \begin{minipage}[ht]{0.15\textwidth}
    \includegraphics[width=\linewidth]{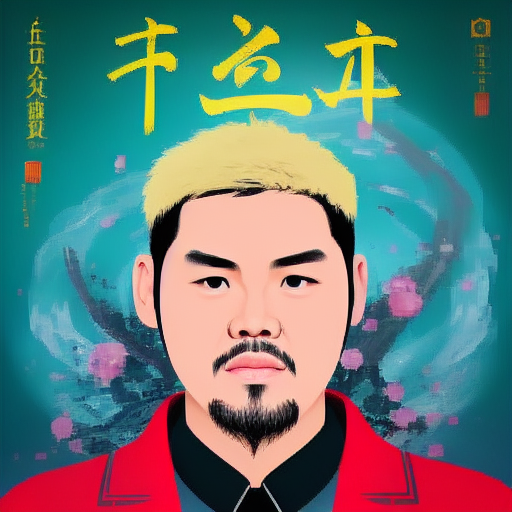}
    \end{minipage}
    \hspace{\fill}
    \begin{minipage}[ht]{0.15\textwidth}
    \includegraphics[width=\linewidth]{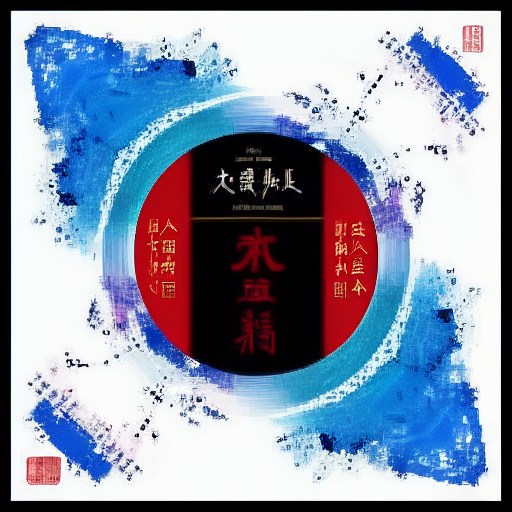}
    \end{minipage}
    \hspace{\fill}
    \begin{minipage}[ht]{0.15\textwidth}
    \includegraphics[width=\linewidth]{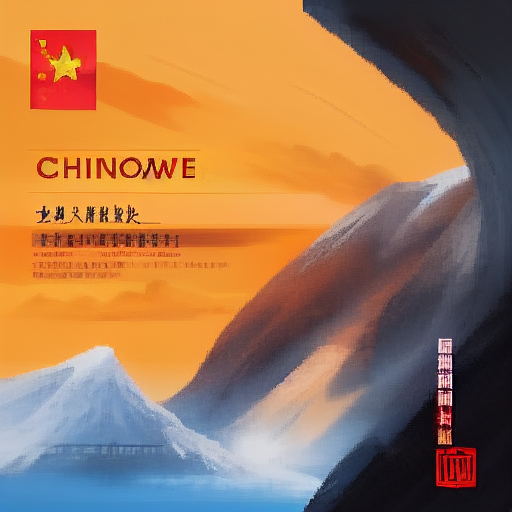}
    \end{minipage}

    \vspace*{0.2cm} 

    \begin{minipage}[ht]{0.15\textwidth}
    \includegraphics[width=\linewidth]{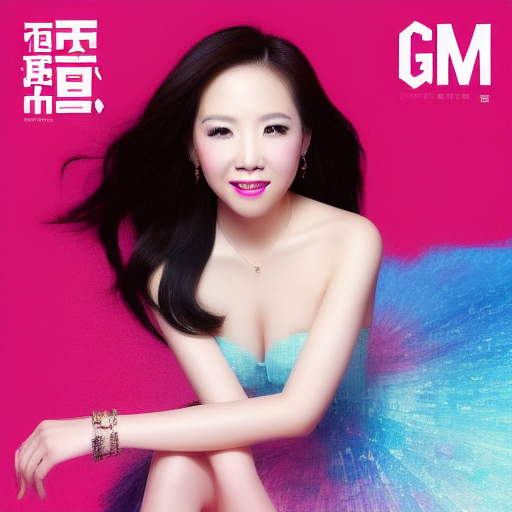}
    \end{minipage}
    \hspace{\fill}
    \begin{minipage}[ht]{0.15\textwidth}
    \includegraphics[width=\linewidth]{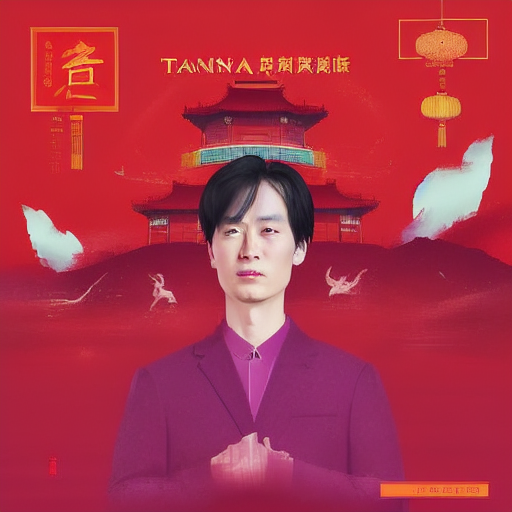}
    \end{minipage}
    \hspace{\fill}
    \begin{minipage}[ht]{0.15\textwidth}
    \includegraphics[width=\linewidth]{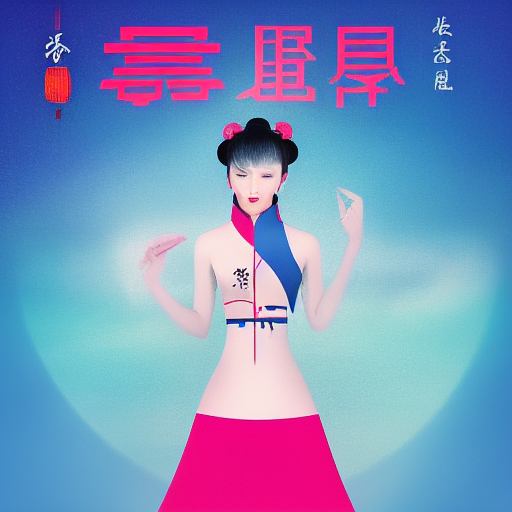}
    \end{minipage}
\caption{Album covers generated by Taiyi diffusion model in the style of popular Chinese artists}
\label{fig:guesswho}
\end{figure}

\section{Experimental Setup}
\label{sec:experimentalsetup}
Inference on BLOOM-176B was performed via the Huggingface Inference API.  All other calculations (stable diffusion, \name, FID) were performed on a machine with an Intel i7-6800K 12-core CPU, 32GB RAM, and a 24GB VRAM RTX 3090 GPU.
Seeds starting at 0 are used throughout our experiments for reproducibility.



\subsection{BLOOM}

In order to use BLOOM to generate lyrics in a zero-shot setting, we provide the following prompt:

\begin{quote}
\centering
\includegraphics[width=\linewidth]{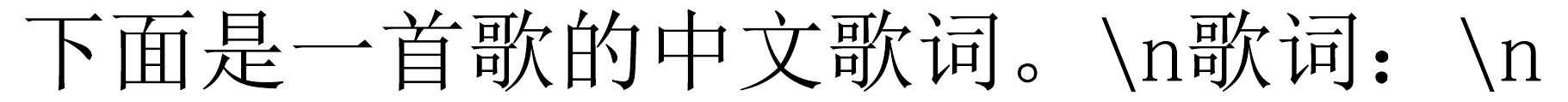}
\scriptsize
Translation: Below are a song's Chinese lyrics. \textbackslash n Lyrics: \textbackslash n
\end{quote}

This prompt was chosen based on the observed result of Chinese language song lyrics.  Shorter prompts such as simply the words for ``lyrics" in Chinese more often resulted in non-song lyrics such as interviews with musicians, and more specific prompts that mentioned specific artists or too closely resembled formatting from specific lyric websites had a higher incidence of generating lyric sequences from real songs. Memorization in large language models is well-documented and is a significant problem undergoing research \cite{carlini2021extracting}.  The selected prompt appears to minimize this behaviour, though future analysis on memorization would be valuable.  

\begin{table}[]
\centering
\scriptsize
\renewcommand*{\arraystretch}{1.4}
\begin{adjustbox}{center}
\begin{tabular}{@{}lllllll@{}}
\toprule
Source         & rep-2 $\downarrow$ & rep-3 $\downarrow$ & rep-4 $\downarrow$ & diversity $\uparrow$ & distinct-2 $\uparrow$ & \name $\uparrow$ \\ \midrule
Human          & \textbf{0.021} &  \textbf{0.013} &  \textbf{0.012} &      \textbf{0.954} &       \textbf{0.870} & -   \\ \hline 
$LLM_{p=0.80}$ & 0.120 &  0.090 &  0.072 &      0.743 &       0.650 & 0.139  \\
$LLM_{p=0.85}$ & 0.095 &  0.073 &  0.063 &      0.786 &       0.689 & 0.162  \\
$LLM_{p=0.90}$ & 0.072 &  0.055 &  0.047 &      0.835 &       0.722 & 0.207  \\
$LLM_{p=0.95}$ & 0.054 &  0.044 &  0.039 &      0.868 &       0.760 & 0.269  \\ 
$LLM_{p=0.99}$ & \textbf{0.040} &  \textbf{0.034} &  \textbf{0.032} &      \textbf{0.897} &       \textbf{0.783} & \textbf{0.293}  \\ \bottomrule
\end{tabular}
\end{adjustbox}
\caption{Quantitative degeneration metrics and \name divergence-based quality results for BLOOM-176B generated text at varying sampling probability mass $p$}
\label{table:bloomy}
\end{table}

\subsection{Amazon Mechanical Turk}

Using Amazon Mechanical Turk to evaluate open-ended text generation is challenging and requires care to perform effectively \cite{karpinska2021perils}, however, such platforms remain useful for their ability to provide real human feedback,  A selection of 60 samples were taken from the BLOOM generated lyric corpora at each value for $top-p$ nucleus sampling, as well as lyrics from the MojimLyrics dataset.  For each of these 60 samples, 3 different reviewers were assigned to score the lyrics for coherence, creativity, affinity (how much the reviewer likes the lyrics), and recognition (whether they believe they may have seen the lyrics before). Answers for each question are converted into normalized scores between 0 and 10.



Instructions were provided in Simplified Chinese.  To encourage close reading, a starting question requests a short summary of the lyrics, and a final multiple-choice question tests the reviewer's basic fluency in Chinese by placing the option ``I can read Chinese" among various other negative options.  Participant responses were removed if they failed to provide a summary or respond correctly to the literacy-testing question.

The projected task duration was 120 seconds, with payment of \$0.44 CAD for a projected hourly wage of \$13.20 CAD.  Actual mean task duration was confirmed to be within the projected task duration.  The survey questions are provided in the GitHub repository.


\subsection{MAUVE Calculation}

For calculating \name, we featurize $p$ with a random sample of 3000 real lyrics from the MojimLyrics dataset, and featurize $q$ with 3000 generated lyrics from BLOOM-176B at each probability mass from our nucleus sampling $p \in [0.80, 0.85, 0.90, 0.95, 0.99]$.  We set a maximum text length of 128 and set a batch size of 32 based on the GPU memory available.  The seed for the \name calculation is set to 0. 






\section{Results}

\begin{figure}[]
\centering
\includegraphics[width=1.0\linewidth]{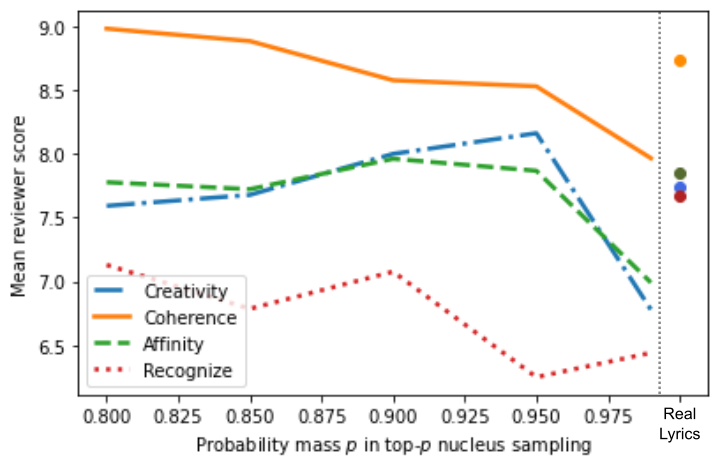}
\caption{Mean human scores for subjective attributes of generated text at varying sampling probability mass $p$. Included to the right are reviewer scores for real lyrics.}
\label{figure:humans}
\end{figure}

Results for language model evaluation with \name scores and degeneration metrics can be found in Table \ref{table:bloomy}, while results of stable diffusion CFG variation on FID can be found in Table \ref{table:sd}.   The complete dataset of generated BLOOM lyrics are provided in the GitHub repository.  Human assessment of lyrics can be found in Figure \ref{figure:humans}.


A sample of generated lyrics and the resulting stable diffusion art can be found in Figure \ref{fig:swordartonlinedieinthegamedieinreallifekekits418am}.  A wider range of stable diffusion images using the album art prompt can be found in Figure \ref{fig:guesswho}, demonstrating the variety of generations the model can produce.  FID scores calculated between outputs of the diffusion model and real album art can be found in Table \ref{table:sd}.


\begin{table}[]
\centering
\renewcommand*{\arraystretch}{1.4}
\begin{tabular}{@{}llll@{}}
\toprule
          & CFG@4.0       &  CFG@7.0 &  CFG@10.0  \\ \midrule
FID$\downarrow$ &   90.21        &   93.84 &  95.91               \\ \bottomrule
\end{tabular}
\caption{FID with real album dataset of Taiyi stable diffusion generated album art at varying CFG}
\label{table:sd}
\end{table}

On the 10 point normalized scale, mean standard deviation among multiple responses for the same input sample were observed to be 1.29, 1.68, 1.95, and 2.67 for coherence, affinity, creativity, and recognition respectively.  However, despite the relatively low variance in response scores, measures of inter-annotator agreement (IAA) are consistently low (i.e., negative Krippendorff ordinal alpha).  Survey responses were highly skewed towards positive answers.  When a reviewer gave a particular sample a low score, often it was only one reviewer did so.  This implies that while reviewers largely agreed on ratings, there was not a consistent criterion across reviewers for giving low scores.  Obtaining high IAA scores on concepts such as creativity and personal preference will likely require a larger study with a greater number of trained reviewers.

\section{Discussion}

Sampling BLOOM for lyrics from a greater portion of the probability mass tends to improve quantitative metrics of generated lyric quality, as shown in Table \ref{table:bloomy}. As expected, token diversity, distinct-$n$, and rep-$n$ all improve at increasing $p$ portion of the probability mass.  That \name score also continues improving implies a low incidence of Type I errors where the model output distribution $Q$ places high probability mass on text which is unlikely under $P$.  Notably, even at very high $p$-values, BLOOM-176B generated lyrics are still more repetitive and less diverse than human lyrics.

Human assessment of lyrics at varying $p$ is charted in Figure \ref{figure:humans}.  Coherence is highest at low-$p$, where common words are frequently sampled, but with a low accompanying creativity score.  Subjective quality measurements of creativity and affinity peak around $p=[0.9, 0.95]$, and trend sharply downwards at $p=0.99$ as the model selects less common words and becomes less coherent.  This implies that for a human to recognize work as ``creative", it must be coherent enough to be understood, but distinctive enough to be interesting.  While \name rises alongside subjective creativity from $p=0.80$ to $p=0.95$, this relationship breaks down at $p=0.99$, implying that \name alone is not a complete replacement for subjective human evaluation.


Finally, it is notable that real popular lyrics uploaded to Mojim.com are at points deemed less creative, coherent, and likable than BLOOM-176B generations.  Resemblance to a human dataset does not inherently bestow positive perceived qualities to generated art.  While the best-scoring CFG setting in Table \ref{table:sd} appears to be at CFG=4.0, this measurement alone is not indicative of subjective human quality.

\section{Conclusion}

In conclusion, we have demonstrated that current computational techniques for evaluating large language model output (both quantitative degeneration metrics and \name), are insufficient as proxies for subjective assessment of creative writing.  Based on human reviewers, we find that large language models such as BLOOM-176B can be used to produce zero-shot song lyrics that are evaluated favourably in comparison to real song lyrics.  Inpsired by AI-driven creativity, we provide a combined approach for generating album art based on lyrics using diffusion. Finally, we provide a novel dataset of in-the-wild popular Chinese song lyrics, and the code for repeating our experiments.

\subsection{Future Work}

We identify the following open problems for future research.

\begin{itemize}
    \item Quantification or representation of abstract concepts such as creativity, and shaping generative models to produce outputs that manifest such attributes, across modalities.
    \item Analysis of sampling parameters and prompts on characteristics of LLM text generation, incorporating both \name and subjective human assessment (including a more comprehensive subjective study).
    \item Deeper study of large multilingual language models such as BLOOM-176B, on topics such as training data memorization, overparameterization, and fairness.
\end{itemize}









\section{Acknowledgments}
Thank you to Annie Feng for survey proofreading, to Bowen Gu for past discussions on lyric generation, and workers at Amazon Mechanical Turk for participating in this study.

\bibliography{aaai23}

\end{document}